\documentclass[letterpaper]{article} 
\usepackage{aaai24}  
\usepackage{times}  
\usepackage{helvet}  
\usepackage{courier}  
\usepackage[hyphens]{url}  
\usepackage{graphicx} 
\usepackage{natbib}  
\usepackage{caption} 
\usepackage{algorithm}
\usepackage{algorithmic}
\usepackage{booktabs}
\usepackage{multirow}
\usepackage{rotating}
\usepackage{amssymb}
\usepackage{amsmath, bm}

\usepackage{amsfonts} 

\usepackage{newfloat}
\usepackage{listings}
\DeclareCaptionStyle{ruled}{labelfont=normalfont,labelsep=colon,strut=off} 
\floatstyle{ruled}
\newfloat{listing}{tb}{lst}{}
\floatname{listing}{Listing}
\title{TimesURL: Self-supervised Contrastive Learning for Universal Time Series Representation Learning}
\author{
    Jiexi Liu,
    Songcan Chen\thanks{Corresponding Author}
}
\affiliations{
    \textsuperscript{\rm 1}College of Computer Science and Technology, Nanjing University of Aeronautics and Astronautics\\
    \textsuperscript{\rm 2}MIIT Key Laboratory of Pattern Analysis and Machine Intelligence\\    

    \{liujiexi, s.chen\}@nuaa.edu.cn
}

\usepackage{bibentry}

\begin{document}

\maketitle

\begin{abstract}
Learning universal time series representations applicable to various types of downstream tasks is challenging but valuable in real applications. Recently, researchers have attempted to leverage the success of self-supervised contrastive learning (SSCL) in Computer Vision(CV) and Natural Language Processing(NLP) to tackle time series representation. Nevertheless, due to the special temporal characteristics, relying solely on empirical guidance from other domains may be ineffective for time series and difficult to adapt to multiple downstream tasks. To this end, we review three parts involved in SSCL including 1) designing augmentation methods for positive pairs, 2) constructing (hard) negative pairs, and 3) designing SSCL loss. For 1) and 2), we find that unsuitable positive and negative pair construction may introduce inappropriate inductive biases, which neither preserve temporal properties nor provide sufficient discriminative features. For 3), just exploring segment- or instance-level semantics information is not enough for learning universal representation. To remedy the above issues, we propose a novel self-supervised framework named TimesURL. Specifically, we first introduce a frequency-temporal-based augmentation to keep the temporal property unchanged. And then, we construct double Universums as a special kind of hard negative to guide better contrastive learning. Additionally, we introduce time reconstruction as a joint optimization objective with contrastive learning to capture both segment-level and instance-level information. As a result, TimesURL can learn high-quality universal representations and achieve state-of-the-art performance in 6 different downstream tasks, including short- and long-term forecasting, imputation, classification, anomaly detection and transfer learning.
\end{abstract}

\begin{figure*}[t]
    \centering
    \includegraphics[width=0.9\textwidth]{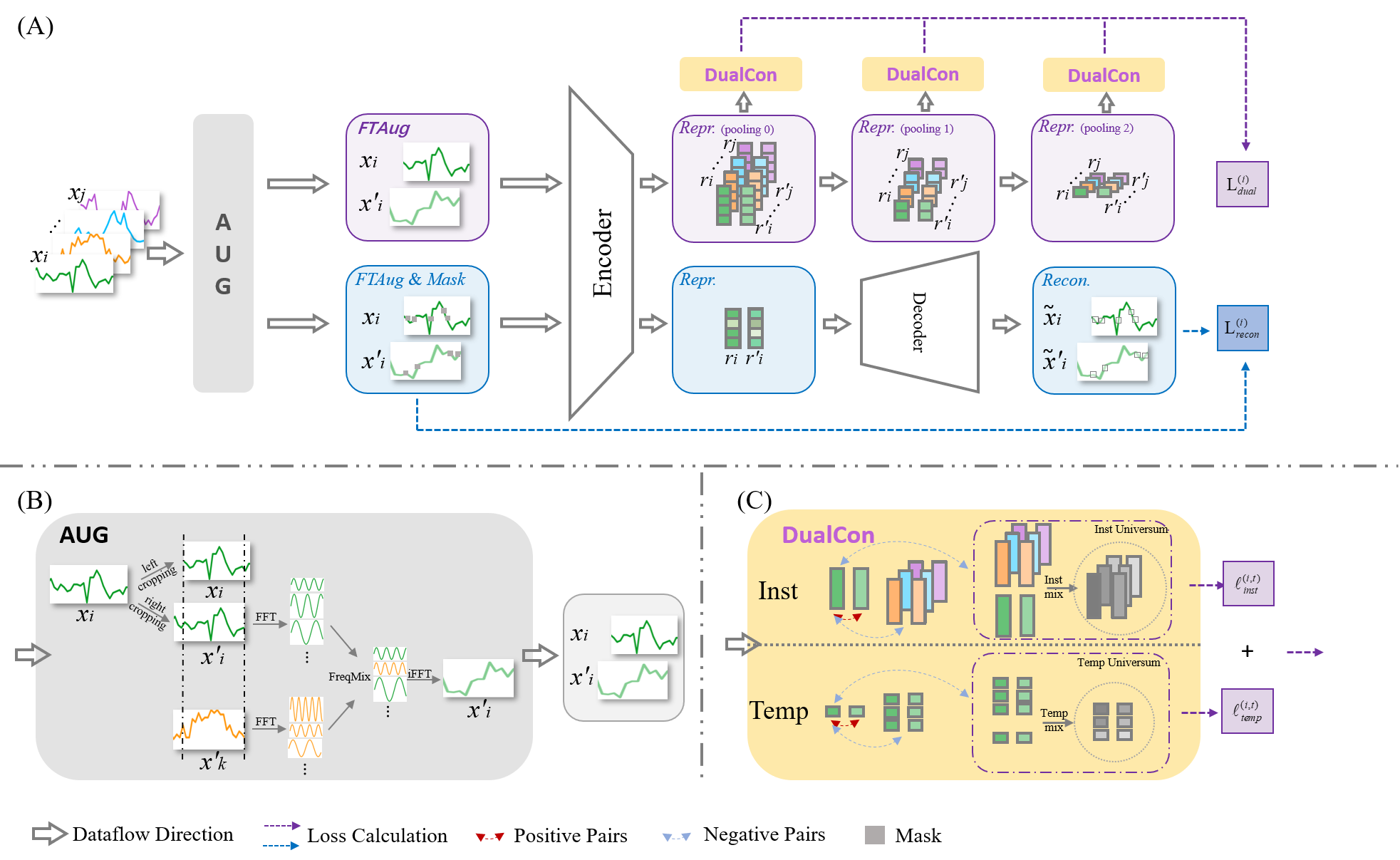}
    \caption{Overview of TimesURL approach, shown in (A), consisting of FTAug, DualCon, and Recon three components. A time series $x_i$ is transformed into two augmented series $x_i$ and $x'_i$ by cropping and frequency mix. Then, the corresponding representation $r_i$ and $r'_i$, the colorful pieces in the rectangular box marked by Repr., are extracted by the Encoder. Within each pooling, shown in light purple rectangular boxes, the learned representations are fed into the DualCon component to synthesize the temporal- and instance-wise Universums, thereby injecting them into contrastive learning. The light blue rectangular boxes represent the reconstruction data flow. Subfigures (B) and (C) denote the specific process of FTAug and Universum synthetic.}
    \label{fig:timesurl}
\end{figure*}

\section{Introduction}
\label{Introduction}
Time series data is ubiquitous in reality ranging from weather and economics to transportation \cite{wu2021autoformer, liu2022non, shi2015convolutional}. Learning information-rich and universal time series representations for multi-type downstream tasks is a fundamental but unsolved problem. While self-supervised contrastive learning has exhibited great success in computer vision (CV), natural language processing (NLP), and recently, other types of modalities \cite{denton2017unsupervised, gutmann2012noise, wang2015unsupervised, pagliardini2017unsupervised, chen2020simple},  its application to time series requires tailored solutions. This is due to the high dimensionality and special temporal characteristics of time series data, as well as the need for diverse semantics information for different tasks. 
 
To this end, we review the four main parts involved in SSCL including 1) augmentation method for positive samples designing, 2) backbone encoder, 3) (hard) negative pairs, and 4) SSCL loss for pretext tasks, and try to invest efforts to explore more effective solutions for time series feature capturing in universal representation learning. Since the backbone encoder has been extensively studied in time series encoder learning \cite{liu2019non, zhou2021informer,wu2023timesnet,liu2022scinet}, our attention is primarily directed toward the remaining three components.

First, most augmentation methods, when applied to time series data, may introduce inappropriate inductive biases as they directly borrow ideas from the fields of CV and NLP. For example, \textit{Flipping} \cite{luo2023time} flip the sign of the original time series that assumes the time series has symmetry between up and down directions. Nevertheless, this may ruin the temporal variations, such as trend, and peak valley, that are inherently present in the original time series. While \textit{permutation} \cite{um2017data} rearranges the order of segments in a time series to generate a new series, under the assumption that the underlying semantic information remains unchanged by the different orders. However, this disturbs the temporal dependencies, thereby impacting the relationships between past and future timestamp information. Consequently, since valuable semantic information of time series primarily resides in temporal variations and dependencies, such augmentations are unable to capture the appropriate features necessary for effective universal representation learning.

Then, the importance of hard negative sample selection has been proved in other domains \cite{kalantidis2020hard, robinson2020contrastive}, but is still underexplored in time series literature. Due to the local smoothness and Markov property, most time series segments can be considered as easy negative samples. These segments tend to exhibit semantic dissimilarity with the anchor and contribute only minor gradients, thus failing to provide useful discriminative information \cite{Cai2020AreAN}. Although the inclusion of a small number of hard negative samples, which have similar but not identical semantics to the anchor, has shown to facilitate improved and expedited learning \cite{xu2022negative, Cai2020AreAN}, their effectiveness is overshadowed by the abundance of easy negative samples.

Last but not least, only using information at segment- or instance-level alone is not enough for learning a universal representation. Prior research has generally classified the aforementioned tasks into two categories \cite{yue2022ts2vec}. The first category includes forecasting, anomaly detection, and imputation that rely more on fine-grained information captured in segment level \cite{yue2022ts2vec, woo2022cost, luo2023time} as these tasks require inferring specific timestamps or sub-sequences. While the second category consists of classification and clustering that prioritize instance-level information, i.e. coarse-grained information \cite{Eldele2021TimeSeriesRL, emadeldeen2022catcc, Liu2022TheTP}, aiming to infer the target across the entire series. Therefore, when confronted with a task-agnostic pre-training model that lacks prior knowledge or awareness of specific tasks during the pre-training phase, both segment- and instance-level information become indispensable for achieving effective universal time series representation learning.

To address these challenges, in this paper, we propose a novel self-supervised framework termed \textbf{TimesURL} to learn universal representations capable of effectively supporting various downstream tasks. We first conduct instance-wise and temporal contrastive learning to incorporate temporal variations and sample diversity. Specifically, to maintain the temporal variations and dependencies, we design a new frequency-temporal-based augmentation method called FTAug which is a combination of cropping in time domain and frequency mixing in the frequency domain. Moreover, inspired by the concept of learning through contradiction, we elaborately design double Universums as hard negative samples. It is a kind of anchor-specific mixup in the embedding space that mixup the specific positive sample (anchor) each time with a negative sample. Our designed double Universums are generated on instance-wise and temporal dimensions respectively, serving as special high-quality hard negative samples that boost the performance of contrastive learning. Additionally, we observe that contrastive learning alone is limited to capturing only one level of information. Therefore, in our paper, we jointly optimize contrastive learning and time reconstruction to capture and leverage information at both segment- and instance levels.

Benefitting from the aforementioned designs, TimesURL consistently achieves state-of-the-art (SOTA) results across a broad range of downstream tasks, thereby demonstrating its ability to learn universal and high-quality representations for time series data. The contributions of this work can be summarized as follows:
\begin{itemize}
\item We revisit the existing contrastive learning framework for time series representation and propose TimesURL, a novel framework that can capture both segment- and instance-level information for universal representation with an additional time reconstruction module.
\item We introduce a new frequency-temporal-based augmentation method and inject novel double Universums into contrastive learning to remedy the positive and negative pairs construction problems. 
\item We evaluate the performance of representation learned by TimesURL via 6 benchmark time series tasks with about 15 baselines. The consistent SOTA performance proves the universality of representation.
\end{itemize}

\section{Related Work}
\paragraph{Unsupervised Representation Learning for Time Series.}
Representation learning for time series has been well-studied for years \cite{chung2015recurrent, krishnan2017structured,bayer2020mind}. 
However, there is still a dearth of research focusing on the more challenging aspect of unsupervised representation learning. SPIRAL \cite{lei2017similarity} bridges the gap between time series data and static clustering algorithms by learning a feature representation that effectively preserves the pairwise similarities inherent in the raw time series data. TimeNet \cite{malhotra2017timenet} is a recurrent neural network that trains the encoder-decoder pair to minimize the reconstruction error from its learned representations. DTCR \cite{ma2019learning} integrates the temporal reconstruction and K-means objective into the seq2seq model to learn cluster-specific temporal representations. ROCKET \cite{dempster2020rocket} is a classification method with small computational expense and fast speed that transforms time series using random convolutional kernels and uses the transformed features to train a linear classifier. Therefore, numerous previous studies concentrate on developing encoder-decoder architectures to minimize reconstruction errors for unsupervised time series representation learning. Some \cite{lei2017similarity,ma2019learning} have attempted to leverage the inherent correlations present in time series data, but have fallen short of fully realizing time series data potential.

\paragraph{Time-Series Contrastive Learning.}
Self-supervised contrastive learning intends to learn invariant representations from different augmented views of data. It is another type of representation learning method for unannotated data over designed pretext tasks. TS-TCC \cite{Eldele2021TimeSeriesRL} focuses on designing a challenging pretext task for robust representation learning from time series data. It tackles this by designing a tough cross-view prediction task against perturbations introduced by different timestamps and augmentations. TNC \cite{tonekaboni2021unsupervised} discusses the choice of positive and negative pair construction by a novel neighborhood-based method for nonstationary multivariate time series with sample weight adjustment. InfoTS \cite{luo2023time} highlights the importance of selecting appropriate augmentations and designs an automatically selecting augmentation method with meta-learning to prevent introducing prefabricated knowledge. TS2Vec\cite{yue2022ts2vec} is a unified framework that learns contextual representations for arbitrary sub-series at various semantic levels. CoST\cite{woo2022cost} contributes to pretext task design by leveraging inductive biases in the model architecture. It specifically focuses on learning disentangled seasonal and trend representations and incorporates a novel frequency domain contrastive loss to encourage discriminative seasonal representations. However, they are prone to be affected by improper prior assumptions, an overabundance of easy negative samples, and a lack of sufficient information for downstream tasks. These limitations arise from inappropriate augmentation methods, the lack of hard negative samples, and the neglect of leveraging both segment- and instance-level information. In this paper, we address all these problems in a unified framework for universal representation learning for time series.

\section{Proposed TimesURL Framework}
\label{section2}
In this section, we make an elaborate description of the newly designed frame, TimesURL. We first formulate the representation learning problem and subsequently delve into the implementation of the key components including contrastive learning and time reconstruction. Particularly, within the contrastive learning component, we emphasize our designed augmentation and double Universum synthesizing methods.

\paragraph{Problem Formulation.}
Similar to most time series representation learning methods, our goal is to learn a nonlinear embedding function $f_\theta$, such that each instance $ {x}_i$ in time series set $ {\mathcal{X}} = \{ {x}_1,{x}_2,\ldots, {x}_N\}$ can map to the best described representation $ {r}_i$. Each input time series instance is $ {x}_i \in \mathbb{R}^{T\times F}$, where $T$ is the time series length and $F$ is the feature dimension. The representation for the $i$-th time series is ${r}_i = \{r_{i,1},r_{i,2},\ldots,r_{i,T}\}$, in which $r_{i,t}\in \mathbb{R}^K$ is the representation vector at time $t$, where $K$ is the dimension of representation vector. Since our model is a two-step progress, we then use the learned representation to accomplish the downstream tasks. 

\paragraph{Method Introduction.}
As shown in Figure \ref{fig:timesurl}, we first generate augmentation sets $\mathcal{X}'$ and $\mathcal{X}_M'$ through FTAug for original series $\mathcal{X}$ and masked series $\mathcal{X}_M$, respectively. Then we get two pairs of original and augmentation series sets, the first pair $(\mathcal{X}, \mathcal{X}')$ is for contrastive learning, while the second pair $(\mathcal{X}_M, \mathcal{X}'_M)$ is for time reconstruction. After that, we map the above sets with $f_\theta$ to achieve corresponding representations. We encourage $\mathcal{R}$ and $\mathcal{R}'$ to have transformation consistency and design a reconstruction method to precisely recover the original dataset $\mathcal{X}$ using both $\mathcal{R}_M$ and $\mathcal{R}'_M$.

The effectiveness of the model above is guaranteed by 1) using a suitable augmentation method for positive pair construction, 2) having a certain amount of hard negative samples for model generalization, and 3) optimizing the encoder $f_\theta$ by contrastive learning and time reconstruction losses jointly for capturing both levels of information. We will then discuss the three parts in the following subsections. 

\subsection{FTAug Method}

A key component of contrastive learning is to choose appropriate augmentations that can impose some priors to construct feasible positive samples so that encoders can be trained to learn robust and discriminative representations \cite{chen2020simple, grill2020bootstrap, yue2022ts2vec}. Most augmentation strategies are task-dependent \cite{luo2023time} and may introduce strong assumptions of data distribution. More seriously, they may perturb the temporal relationship and semantic consistency that is crucial for tasks like forecasting. Therefore, we choose the contextual consistency strategy \cite{yue2022ts2vec}, which treats the representations at the same timestamp in two augmented contexts as positive pairs. Our FTAug combines the advantages in both frequency and temporal domains that generate the augmented contexts by frequency mixing and random cropping.

\paragraph{Frequency mixing.} Frequency mixing is used to produce a new context view by replacing a certain rate of the frequency components in one training instance $x_i$ calculated by Fast Fourier Transform (FFT) operation with the same frequency components of another random training instance $x_k$ in the same batch. Then we use the inverse FFT to convert back to get a new time domain time series \cite{chen2023fraug} Exchanging frequency components between samples will not introduce unexpected noise or artificial periodicities and can offer more reliable augmentations for preserving the semantic characteristics of the data.

\paragraph{Random cropping.} Random cropping is the key step for contextual consistency strategy. For each instance $x_i$, we randomly sample two overlapping time segments $[a_1, b_1]$, $[a_2,b_2]$ such that $0<a_1\le a_2\le b_1\le b_2 \le T$. The contrastive learning and time reconstruction further optimize the representation in the overlapping segment $[a_2, b_1]$.

 Ultimately, the proposed FTAug is helpful for various kinds of tasks since it can keep the important temporal relationship and semantic consistency for time series. We have to mention that the FTAug is only applied in the training process. 

\subsection{Double Universum Learning}
\begin{figure}[t]
    \centering
    \includegraphics[width=0.9\linewidth]{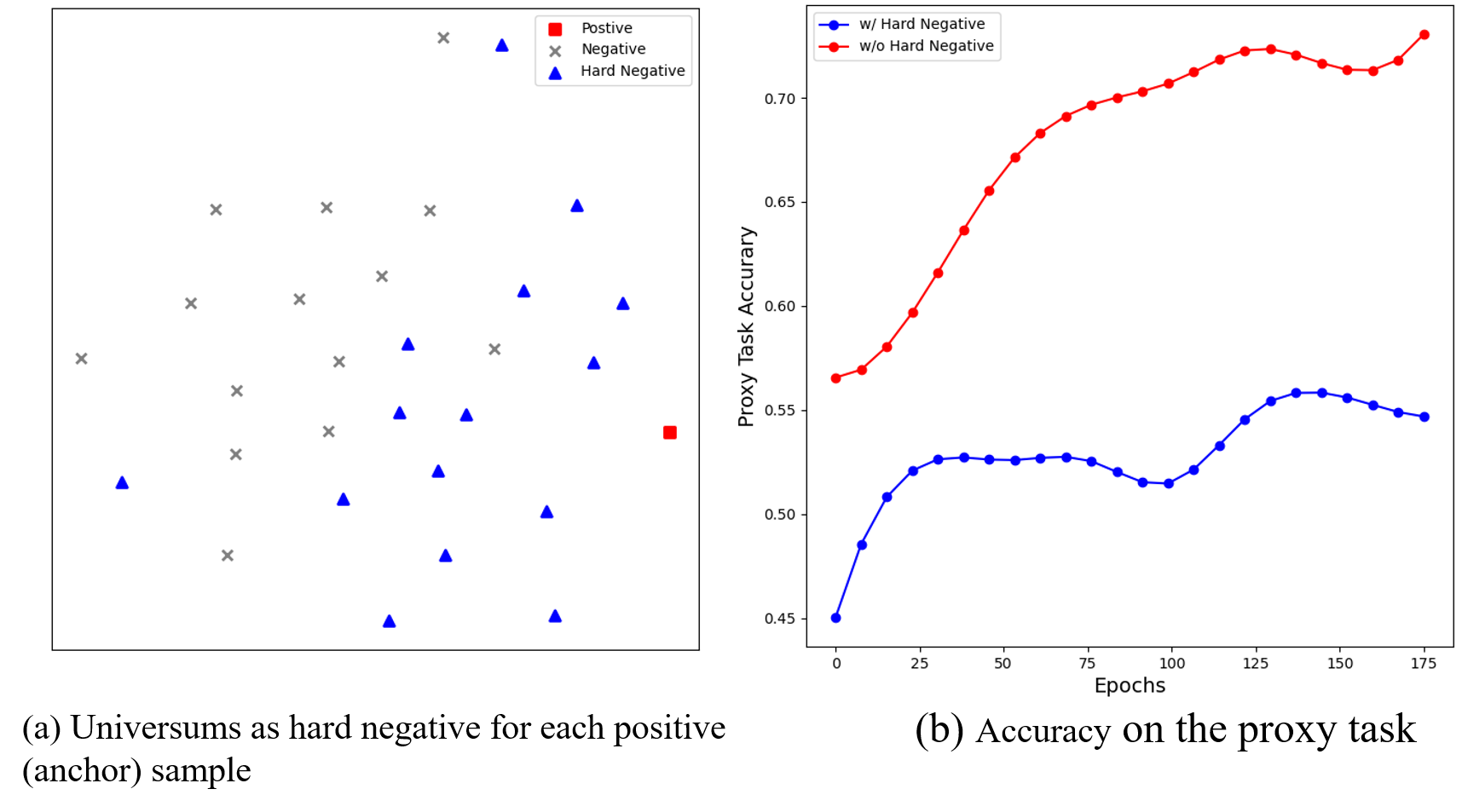}
    \caption{Properties of Universums on ERing dataset in UEA archive}
    \label{fig:uni}
\end{figure}

As revealed by recent studies \cite{kalantidis2020hard, robinson2020contrastive, Cai2020AreAN}, hard negative samples play an important role in contrastive learning but have never been explored in the time series domain. 
Moreover, due to the local smoothness and the Markov property in time series, most negative samples are easy negative samples that are insufficient for capturing temporal-wise information since they fundamentally lack the learning signals required to drive contrastive learning. As a real example of ERing dataset in UEA archive \cite{bagnall2018uea} in Figure \ref{fig:uni}, for each positive anchor (red square), the corresponding negative samples (gray marks) contain many easy negatives and few hard ones, i.e. many of the negatives are too far to contribute to the contrastive loss. 

 Our double Universums are Mixup Induced Universums \cite{han2023universum,vapnik2006Transductive, chapelle2007analysis} in both instance- and temporal-wise, which is anchor-specific mixing in the embedding space that mixes the specific positive feature (anchor) with the negative features for unannotated datasets. 

 Let $i$ be the index of the input time series sample and $t$ be the timestamp. $r_{i,t}$ and $r'_{i,t}$ denote the representations for the same timestamp $t$ but from two augmentation of $x_i$. The synthetic temporal-wise Universums for the $i$-th time series at timestamp $t$ can be formulated as
 \begin{equation}
     \begin{array}{c}
          {r}^{\text{temp}}_{i,t} = \lambda_1 \cdot r_{i,t} + (1 - \lambda_1) \cdot r_{i,t'}, \\
          {r}'^{\text{temp}}_{i,t} = \lambda_1 \cdot r'_{i,t} + (1 - \lambda_1) \cdot r'_{i,t'}, 
     \end{array}
     \label{t-universum}
 \end{equation}
in which $t'$ is randomly chosen from $\Omega$, the set of timestamps within the overlap of the two subseries, and $t' \neq t$. While the instance-wise Universums indexed with $(i,t)$ are similar be formulated as
 \begin{equation}
     \begin{array}{c}
          {r}^{\text{inst}}_{i,t} = \lambda_2 \cdot r_{i,t} + (1 - \lambda_2) \cdot r_{j,t}, \\
          {r}'^{\text{inst}}_{i,t} = \lambda_2 \cdot r'_{i,t} + (1 - \lambda_2) \cdot r'_{j,t},
     \end{array}
     \label{i-universum}
 \end{equation}
 where $j$ indicates any other instance except $i$ in batch $\mathcal{B}$. Here, $\lambda_1,\lambda_2 \in (0,0.5]$ are randomly chosen mixing coefficients for the anchor, and $\lambda_1,\lambda_2 \leq 0.5$ guarantees that the anchor’s contribution is always smaller than negative samples.

As in Figure \ref{fig:uni}(a), most Universum (blue triangles) are much closer to the anchor and thus can be seen as hard negative samples. Moreover, we utilize a proxy task to indicate the difficulty of hard negatives \cite{kalantidis2020hard}, i.e. Universums. The proxy task performance is shown in Figure \ref{fig:uni}(b), i.e. the percentage of anchors where the positive sample is ranked overall negatives across training our TimesURL with and without Universums on ERing dataset. Despite the drop in proxy task performance of TimesURL, however, further performance gains are observed for linear classification from 0.896 (without Universums) to 0.985 (with Universums), which means that the additional Universum makes the proxy task harder to solve but can further improve the model performance in the downstream task. Therefore, Univerums in TimesURL can be seen as high-quality negatives. To sum up, our Universums can be treated as a kind of high-quality hard negative samples.

By mixing with the anchor sample, the possibility of the universum data falling into target regions in the data space is minimized, thereby ensuring the hard negativity of Universum. Moreover, the double Universum set contains all other negative samples that are beneficial to learn discriminative sample information to increase model capability.

\subsection{Contrastive Learning for Segment-level Information}
We use a straightforward way to inject the double Universums into contrastive learning as additional hard negative samples in both temporal- and instance-wise contrastive loss, respectively. The two losses for the $i$-th time series at timestamp $t$ can be formulated as 
\begin{equation}
    \ell_{\text{temp}}^{(i, t)}=-\log \frac{\exp (r_{i, t} \cdot r_{i, t}^{\prime})}
    {\exp{\left(r_{i, t}\cdot r'_{i, t}\right)} + \sum \limits_{z_{i, t'} \in \mathbb{N}_i}{\exp{\left(r_{i, t}\cdot z_{i, t'}\right)}}}
\label{temploss}
\end{equation}
\begin{equation}
    \ell_{\text{inst}}^{(i, t)}=-\log \frac{\exp (r_{i, t} \cdot r_{i, t}^{\prime})}
    {\exp (r_{i, t} \cdot r_{i, t}^{\prime}) + \sum \limits_{z_{j, t} \in \mathbb{N}_j}{\exp{\left(r_{i, t}\cdot z_{j, t}\right)}}}
\label{instloss}
\end{equation}
where in Eq.\eqref{temploss} and \eqref{instloss}, $\mathbb{N}_i \triangleq \mathbb{Z}_i \cup \mathbb{Z}'_i \cup \mathbb{U}_i \cup \mathbb{U}'_i$, and $\mathbb{N}_j \triangleq \mathbb{Z}_j \cup \mathbb{Z}'_j \cup \mathbb{U}_j \cup \mathbb{U}'_j$, in which $\mathbb{Z}_i \cup \mathbb{Z}'_i = \{r_{i,t'}, r'_{i,t'}| t'\in \Omega \backslash t \}$ and $\mathbb{Z}_j \cup \mathbb{Z}'_j = \{r_{j,t}, r'_{j,t}| j \in \{1,\ldots, |\mathcal{B}| \}\backslash i \}$ are original negative samples,  while $\mathbb{U}_i \cup \mathbb{U}'_i = \{{r}^{\text{temp}}_{i,t'},{r}'^{\text{temp}}_{i,t'}|t'\in \Omega \}$ and $\mathbb{U}_j \cup \mathbb{U}'_j = \{{r}^{\text{inst}}_{j,t},{r}'^{\text{inst}}_{j,t}|j \in \{1,\ldots, |\mathcal{B}|\}$ are proposed double Universums as in Eq.\eqref{t-universum},\eqref{i-universum}, in which $|\mathcal{B}|$ denotes the batch size.

The two losses are complementary to each other to capture both instance-specific characteristics and the temporal variation. We use hierarchical contrastive loss \cite{yue2022ts2vec} for multi-scale information learning by using max pooling on the learned representations along the time axis in Eq.\eqref{temploss} and \eqref{instloss}. Here, we have to mention that important temporal variation information, such as trend and seasonal are lost after several max pooling operations, therefore contrasting at top levels cannot actually capture sufficient instance-level information for downstream tasks.

\vspace*{-3pt}
\begin{equation}
\mathcal{L}_{\text{dual}}=\frac{1}{|\mathcal{B}| T} \sum_{i} \sum_{t}\left(\ell_{\text {temp }}^{(i, t)}+\ell_{\text {inst }}^{(i, t)}\right)
\label{dualloss}
\end{equation}

\subsection{Time Reconstruction for Instance-level Information}
The masked autoencoding technique in self-supervised learning has been proven to perform well in various domains, such as BERT-based pre-training model\cite{kenton2019bert} in NLP as well as MAE \cite{he2022masked} in CV. The main idea of such methods is to reconstruct the original signal given its partial observation.

Motivated by the masked autoencoding technique, we design a reconstruction module to preserve important temporal variation information. Our approach uses the above mentioned embedding function $f_\theta$ as an encoder that maps the masked instance into latent representation and then reconstructs the full instance from the latent representation. Here, we use the random masking strategy. Our loss function computes the Mean Squared Error (MSE) between the reconstructed and the original value at each timestamp. Further, similar to BERT and MAE, we compute the MSE loss only on the masked timestamps in Eq.\eqref{reconloss}.
\vspace*{-3pt}
\begin{equation}
\mathcal{L}_{\text {recon}}= \frac{1}{2|\mathcal{B}| } \sum_{i}{\|m_i \odot(\tilde{x}_i-{x}_i)\|_{2}^{2} + \|m'_i \odot(\tilde{x}'_i-{x}'_i)\|_{2}^{2}}
\label{reconloss}
\end{equation}
Here, we denote $m_i \in \{0,1\}^{T\times F}$ as the observation mask for the $i$-th instance where $m_{i,t}=0$ if $x_{i,t}$ is missing, and $m_{i,t}=1$ if $x_{i,t}$ is observed, while $\tilde{x}_i$ is the generated reconstruction instance. Similar to the above notations, $m'_i$, $\tilde{x}'_i$ and ${x}'_i$ have the same meaning.

The overall loss is defined as
\vspace*{-3pt}
\begin{equation}
\mathcal{L}= \mathcal{L}_{\text{dual}} + \alpha \mathcal{L}_{\text{recon}}
\label{dualloss}
\end{equation}
where $\alpha$ is a hyper-parameter to balance the two losses.

\section{Experiments}
In this section, to evaluate the generality and the downstream tasks performance of the representation learned by our TimesURL, we extensively experiment on 6 downstream tasks, including short- and long-term forecasting, imputation, classification, anomaly detection and transfer learning. The best results are highlighted in bold. More detailed experimental setups and other additional experiment results will be presented in the Appendix.

\paragraph{Implementation} The summary of the benchmarks is in the Appendix. For TimesURL, we use  Temporal Convolution Network (TCN) as the backbone encoder, which is similar to TS2Vec \cite{yue2022ts2vec}. More detailed information about the dataset and other experiment implementation information is in the Appendix.

 \begin{table*}[t] 
    \centering
    \resizebox{0.8\textwidth}{!}{
  \begin{tabular}{ccccccccccccc}
  \toprule
    \multicolumn{2}{c}{Method}  & TimesURL & InfoTS & TS2Vec & T-Loss & TNC & TS-TCC & TST & DTW \\
    \midrule
    \multicolumn{1}{c|}{\multirow{2}[2]{*}{30 UEA datasets}}  & \multicolumn{1}{c|}{ Avg. ACC} & \textbf{0.752 (+3.8\%)}& 0.714& 0.704& 0.658 & 0.670 & 0.668 & 0.617 & 0.629 \\
    \multicolumn{1}{c|}{} & \multicolumn{1}{c|}{Avg. Rank} & \textbf{1.367} & 3.200 & 3.567 & 4.567 & 5.333 & 5.000 & 5.900 & 5.207 \\
     \midrule
     \multicolumn{1}{c|}{\multirow{2}[2]{*}{128 UCR datasets}}  & \multicolumn{1}{c|}{ Avg. ACC} & \textbf{0.845 (+0.7\%)} & 0.838 & 0.836 & 0.806 & 0.761 & 0.757 & 0.639 & 0.729\\
    \multicolumn{1}{c|}{} & \multicolumn{1}{c|}{Avg. Rank} & \textbf{1.844} & 2.047 & 2.625 & 4.248 & 5.128 & 5.032 & 6.961 & 6.008 \\
    \bottomrule
  \end{tabular}
}
    \caption{Time series classification results. }\label{tab:full-uea-unsup-sum}
\end{table*}

\paragraph{Baselines} Following the self-supervised learning setting, we extensively compare TimesURL with recent advanced models under a similar experimental setup. Since most existing self-supervised learning methods cannot learn universal representations for all kinds of tasks, we utilize each method only for tasks it is specifically designed for. Moreover, we also compare the SOTA models for each specific task as follows, where SSL, E2EL, USL are abbreviations for self-supervised, end-to-end, and unsupervised learning, respectively: 

\noindent \textit{\textbf{Forecasting}}: 1) \textbf{SSL}: CoST \cite{woo2022cost}, TS2Vec \cite{yue2022ts2vec}, TNC \cite{tonekaboni2021unsupervised}, 2) \textbf{E2EL}: Informer \cite{zhou2021informer}, LogTrans \cite{li2019enhancing}, N-BEATS \cite{oreshkin2019n}, TCN \cite{bai2018empirical};

\noindent \textit{\textbf{Classification}}: 1) \textbf{SSL}: InfoTS \cite{luo2023time}, TS2Vec \cite{yue2022ts2vec}, TNC \cite{tonekaboni2021unsupervised}, TS-TCC \cite{Eldele2021TimeSeriesRL}, TST \cite{zerveas2021transformer},2) \textbf{USL}: DTW;

\noindent \textit{\textbf{Imputation}}: \textbf{SSL}: TS2Vec\cite{yue2022ts2vec}, InfoTS\cite{luo2023time};

\noindent \textit{\textbf{Anomaly detection}}: 1) \textbf{SSL}: TS2Vec\cite{yue2022ts2vec}, 2) \textbf{USL}: SPOT\cite{siffer2017anomaly}, DSPOT\cite{siffer2017anomaly}, DONUT\cite{xu2018unsupervised}, SR\cite{ren2019time}.

 Overall, about 15 baselines are included for a comprehensive comparison.  

\subsection{Classification}
\textbf{Setups}
Time series classification has practical significance in medical diagnosis, action recognition, etc. Our experiments are under the setting that class labels are on the instance. So instance-level classification is adopted to verify the model capacity in presentation learning. We select commonly used UEA \cite{bagnall2018uea} and UCR \cite{dau2019ucr} Classification Archive. The representation dimensions of all classification methods except DTW are set to $320$ and we then follow the same protocol as TS2Vec which uses an SVM classifier with RBF kernel to train on top of representations for classification.

\noindent \textbf{Results}
The evaluation results are shown in Table \ref{tab:full-uea-unsup-sum}. TimesURL achieves the best performance with  an average accuracy of $75.2\%$ for 30 univariate datasets in UEA  and $84.5\%$ 
 for 128 multivariate datasets in UCR, surpassing the previous SOTA self-supervised method InfoTS ($71.4\%$). Moreover, the best average rank also validates the significant outperformance of TimesURL. As mentioned before, it is easy to understand the failure of other methods, since TS2Vec lacks sufficient instance-level information, while some augmentations in InfoTS may introduce inappropriate inductive biases that damage the temporal properties, such as the trend for classification. Moreover, other contrastive learning methods including T-Loss, TS-TCC, TNC and TST perform only at a certain level. Since TimesURL uses general FTAug and contains appropriate hard negatives and can capture both segment- and instance-level information, thus achieves better performance. 

\subsection{Imputation}
\textbf{Setups}
Under a realistic scenario, irregular and asynchronized sampling often happens, which may lead to missingness resulting in difficulty in downstream tasks. Imputation is a straightforward and widely used method to relieve this problem. We complete the task with ETT dataset \cite{zhou2021informer} under the electricity scenario, where the data-missing problem happens commonly. To compare the model capacity under different proportions of missing data, we randomly mask the time points in the ratio of $\{12.5\%, 25\%, 37.5\%, 50\%\}$. We then follow the same setting as TimesNet which uses a MLP network for the downstream tasks.

\begin{table}[h]
    \centering
    \resizebox{0.8\linewidth}{!}{
\begin{tabular}{cccccccc}
\toprule
\multicolumn{2}{c}{} & \multicolumn{2}{c}{TimesURL} & \multicolumn{2}{c}{InfoTS}  & \multicolumn{2}{c}{TS2Vec} \\
\midrule
\multicolumn{2}{c}{Metrics} & MSE   & MAE   & MSE   & MAE   & MSE   & MAE  \\
\midrule
\multicolumn{1}{c|}{\multirow{4}[2]{*}{\begin{sideways}ETTh1\end{sideways}}} & \multicolumn{1}{c|}{0.125} & \textbf{0.659} & \textbf{0.640} & 0.717 & 0.666 & 0.690 & 0.658 \\
\multicolumn{1}{c|}{} & \multicolumn{1}{c|}{0.250} & \textbf{0.679} & \textbf{0.648} & 0.726 & 0.674 & 0.710 & 0.668 \\
\multicolumn{1}{c|}{} & \multicolumn{1}{c|}{0.375} & \textbf{0.702} & \textbf{0.656} & 0.726 & 0.676 & 0.728 & 0.676 \\
\multicolumn{1}{c|}{} & \multicolumn{1}{c|}{0.500} & \textbf{0.712} & 0.693 & 0.783 & 0.695 & 0.751 & \textbf{0.682} \\
\midrule
\multicolumn{1}{c|}{\multirow{4}[2]{*}{\begin{sideways}ETTh2\end{sideways}}} & \multicolumn{1}{c|}{0.125} & \textbf{2.455} & 1.215 & 2.491 & \textbf{1.199} & 2.866 & 1.288 \\
\multicolumn{1}{c|}{} & \multicolumn{1}{c|}{0.250} & \textbf{2.560} & \textbf{1.239} & 2.644 & 1.244 & 2.792 & 1.271 \\
\multicolumn{1}{c|}{} & \multicolumn{1}{c|}{0.375} & \textbf{2.673} & 1.269 & 2.757 & \textbf{1.266} & 2.793 & 1.271 \\
\multicolumn{1}{c|}{} & \multicolumn{1}{c|}{0.500} & \textbf{2.701} & 1.281 & 2.844 & 1.283 & 2.769 & \textbf{1.267} \\
\midrule
\multicolumn{1}{c|}{\multirow{4}[2]{*}{\begin{sideways}ETTm1\end{sideways}}} & \multicolumn{1}{c|}{0.125} & \textbf{0.644} & \textbf{0.650} & 0.702 & 0.651 & 0.726 & 0.663 \\
\multicolumn{1}{c|}{} & \multicolumn{1}{c|}{0.250} & \textbf{0.699} & 0.668 & 0.732 & 0.664 & 0.719 & \textbf{0.664} \\
\multicolumn{1}{c|}{} & \multicolumn{1}{c|}{0.375} & \textbf{0.718} & 0.686 & 0.753 & 0.674 & 0.728 & \textbf{0.670} \\
\multicolumn{1}{c|}{} & \multicolumn{1}{c|}{0.500} & \textbf{0.716} & 0.680 & 0.759 & 0.677 & 0.739 & \textbf{0.669} \\
\midrule
\multicolumn{2}{c}{Avg.} & \textbf{1.326} & \textbf{0.860} & 1.386 & 0.864 & 1.418 & 0.871 \\
\bottomrule
\end{tabular}%

    }
    \caption{Multivarite time series imputation results.}
    \label{tab:imputation}
\end{table}

\noindent \textbf{Results}
Since TimesURL contains a time construction module to capture underlying temporal patterns, we naturally extend it to a downstream task. As shown in Table \ref{tab:imputation}, our proposed TimesNet still achieves SOTA performance on three datasets and proves to have the ability to capture temporal variation from complicated time series.

\subsection{Short- and Long-Term Forecasting}
\textbf{Setups}
Time series forecasting is ubiquitous in our
everyday life. For both short- and long- term forecasting we use ETT, Electricity and Weather datasets from various reality scenarios and the results of the two latter datasets are in the Appendix. For short-term forecasting, the horizon is $24$ and $48$, while for long-term forecasting the horizon ranges from $96$ to $720$. Here, we learn the representations once for each dataset and can be directly applied to various horizons with linear regressions. This helps demonstrate the universality of the learned representation. 

\noindent \textbf{Results}
We compare TimesURL with not only representation learning as well as end-to-end forecasting methods in \ref{tab:forecasting} and indicates that TimesURL has established a new SOTA in most cases for both short- and long-term forecasting. 

\begin{table*}[t]
  \centering
  \resizebox{\textwidth}{!}{
\begin{tabular}{cccccccccccccccccc}
\toprule
\multicolumn{2}{c}{\multirow{2}[4]{*}{Methods}} & \multicolumn{8}{c}{Representation Learning}                   & \multicolumn{8}{c}{End-to-end Forecasting}\\
\cmidrule{3-18}\multicolumn{2}{c}{} & \multicolumn{2}{c}{TimesURL} & \multicolumn{2}{c}{CoST}  & \multicolumn{2}{c}{TS2Vec} & \multicolumn{2}{c}{TNC} & \multicolumn{2}{c}{Informer} & \multicolumn{2}{c}{LogTrans}  & \multicolumn{2}{c}{N-BEATS} & \multicolumn{2}{c}{TCN} \\
\midrule
\multicolumn{2}{c}{Metrics} & MSE   & MAE   & MSE   & MAE   & MSE   & MAE   & MSE   & MAE   & MSE   & MAE   & MSE   & MAE   & MSE   & MAE   & MSE   & MAE \\
\midrule
\multicolumn{1}{c|}{\multirow{5}[2]{*}{\begin{sideways}ETTh1\end{sideways}}} & \multicolumn{1}{c|}{24} & \textbf{0.0355} & \textbf{0.1423} & 0.0400 & 0.1520 & 0.0390 & 0.1510 & 0.0570 & \multicolumn{1}{c|}{0.1840} & 0.0980 & 0.2470 & 0.1030 & 0.2590 & 0.0940 & 0.2380 & 0.1040 & 0.2540 \\
\multicolumn{1}{c|}{} & \multicolumn{1}{c|}{48} & \textbf{0.0535} & \textbf{0.1460} & 0.0600 & 0.1860 & 0.0620 & 0.1890 & 0.0940 & \multicolumn{1}{c|}{0.2390} & 0.1580 & 0.3190 & 0.1670 & 0.3280 & 0.2100 & 0.3670 & 0.2060 & 0.3660 \\
\multicolumn{1}{c|}{} & \multicolumn{1}{c|}{168} & \textbf{0.0956} & \textbf{0.2327} & 0.0970 & 0.2360 & 0.1420 & 0.2910 & 0.1710 & \multicolumn{1}{c|}{0.3290} & 0.1830 & 0.3460 & 0.2070 & 0.3750 & 0.2320 & 0.3910 & 0.4620 & 0.5860 \\
\multicolumn{1}{c|}{} & \multicolumn{1}{c|}{336} & 0.1210 & 0.2672 & \textbf{0.1120} & \textbf{0.2580} & 0.1600 & 0.3160 & 0.1920 & \multicolumn{1}{c|}{0.3570} & 0.2220 & 0.3870 & 0.2300 & 0.3980 & 0.2320 & 0.3880 & 0.4220 & 0.5640 \\
\multicolumn{1}{c|}{} & \multicolumn{1}{c|}{720} & \textbf{0.1453} & 0.3068 & 0.1480 & \textbf{0.3060} & 0.1790 & 0.3450 & 0.2350 & \multicolumn{1}{c|}{0.4080} & 0.2690 & 0.4350 & 0.2730 & 0.4630 & 0.3220 & 0.4900 & 0.4380 & 0.5780 \\
\midrule
\multicolumn{1}{c|}{\multirow{5}[2]{*}{\begin{sideways}ETTh2\end{sideways}}} & \multicolumn{1}{c|}{24} & 0.0834 & 0.2186 & \textbf{0.0790} & \textbf{0.2070} & 0.0910 & 0.2300 & 0.0970 & \multicolumn{1}{c|}{0.2380} & 0.0930 & 0.2400 & 0.1020 & 0.2550 & 0.1980 & 0.3450 & 0.1090 & 0.2510 \\
\multicolumn{1}{c|}{} & \multicolumn{1}{c|}{48} & \textbf{0.1158} & \textbf{0.2186} & 0.1180 & 0.2590 & 0.1240 & 0.2740 & 0.1310 & \multicolumn{1}{c|}{0.2810} & 0.1550 & 0.3140 & 0.1690 & 0.3480 & 0.2340 & 0.3860 & 0.1470 & 0.3020 \\
\multicolumn{1}{c|}{} & \multicolumn{1}{c|}{168} & \textbf{0.1747} & \textbf{0.3324} & 0.1890 & 0.3390 & 0.1980 & 0.3550 & 0.1970 & \multicolumn{1}{c|}{0.3540} & 0.2320 & 0.3890 & 0.2460 & 0.4220 & 0.3310 & 0.4530 & 0.2090 & 0.3660 \\
\multicolumn{1}{c|}{} & \multicolumn{1}{c|}{336} & \textbf{0.1875} & \textbf{0.3469} & 0.2060 & 0.3600 & 0.2050 & 0.3640 & 0.2070 & \multicolumn{1}{c|}{0.3660} & 0.2630 & 0.4170 & 0.2670 & 0.4370 & 0.4310 & 0.5080 & 0.2370 & 0.3910 \\
\multicolumn{1}{c|}{} & \multicolumn{1}{c|}{720} & \textbf{0.1862} & \textbf{0.3519} & 0.2140 & 0.3710 & 0.2080 & 0.3710 & 0.2070 & \multicolumn{1}{c|}{0.3700} & 0.2770 & 0.4310 & 0.3030 & 0.4930 & 0.4370 & 0.5170 & 0.2000 & 0.3670 \\
\midrule
\multicolumn{1}{c|}{\multirow{5}[2]{*}{\begin{sideways}ETTm1\end{sideways}}} & \multicolumn{1}{c|}{24} & \textbf{0.0128} & \textbf{0.0839} & 0.0150 & 0.0880 & 0.0160 & 0.0930 & 0.0190 & \multicolumn{1}{c|}{0.1030} & 0.0300 & 0.1370 & 0.0650 & 0.2020 & 0.0540 & 0.1840 & 0.0270 & 0.1270 \\
\multicolumn{1}{c|}{} & \multicolumn{1}{c|}{48} & \textbf{0.0242} & 0.1765 & 0.0250 & \textbf{0.1170} & 0.0280 & 0.1260 & 0.0360 & \multicolumn{1}{c|}{0.1420} & 0.0690 & 0.2030 & 0.0780 & 0.2200 & 0.1900 & 0.3610 & 0.0400 & 0.1540 \\
\multicolumn{1}{c|}{} & \multicolumn{1}{c|}{96} & \textbf{0.0366} & \textbf{0.1445} & 0.0380 & 0.1470 & 0.0450 & 0.1620 & 0.0540 & \multicolumn{1}{c|}{0.1780} & 0.1940 & 0.3720 & 0.1990 & 0.3860 & 0.1830 & 0.3530 & 0.0970 & 0.2460 \\
\multicolumn{1}{c|}{} & \multicolumn{1}{c|}{288} & 0.0797 & 0.2139 & \textbf{0.0770} & \textbf{0.2090} & 0.0950 & 0.2350 & 0.0980 & \multicolumn{1}{c|}{0.2440} & 0.4010 & 0.5540 & 0.4110 & 0.5720 & 0.1860 & 0.3620 & 0.3050 & 0.4550 \\
\multicolumn{1}{c|}{} & \multicolumn{1}{c|}{672} & 0.1141 & \textbf{0.2552} & \textbf{0.1130} & 0.2570 & 0.1420 & 0.2900 & 0.1360 & \multicolumn{1}{c|}{0.2900} & 0.5120 & 0.6440 & 0.5980 & 0.7020 & 0.1970 & 0.3680 & 0.4450 & 0.5760 \\
\midrule
\multicolumn{2}{c}{Avg.} & \textbf{0.0977} & \textbf{0.2292} & 0.1021 & 0.2328 & 0.1156 & 0.2528 & 0.1287 & \multicolumn{1}{c|}{0.2722} & 0.2104 & 0.3623 & 0.2279 & 0.3907 & 0.2354 & 0.3807 & 0.2299 & 0.3722 \\
\bottomrule
\end{tabular}%
    }
  \caption{Short- and Long-Term Forecasting Univariate forecasting results.}
  \label{tab:forecasting}%
\end{table*}

\subsection{Anomaly Detection}
\textbf{Setups}
Detecting anomalies from monitoring data is essential for industrial maintenance. We follow the setting of a streaming evaluation protocol \cite{ren2019time} in time series anomaly detection that determines whether the last point $x_t$ in time series slice $x_1,\ldots,x_t$ is an anomaly or not. During training, each time series sample is split into two halves according to the time order, where the first half is for training and the second is for evaluation. In this task, We compare models on two benchmark datasets, including KPI \cite{ren2019time} a competition dataset that includes multiple minutely sampled real KPI curves and Yahoo \cite{yahoo} including $367$ hourly sampled time series.

\noindent \textbf{Results}
Table \ref{tab:anomaly} shows the performance of anomaly detection tasks with different methods on F1 score, precision and recall. In the normal setting, TimesURL has consistently good performance on both KPI and Yahoo datasets. 

\subsection{Transfer Learning}
We complete the transfer learning task to demonstrate that the representation learned by TimesURL has good transferability that can achieve good performance when training on one condition (i.e., source domain) and testing it on other multiple conditions (i.e., target domains). Here, we present the transfer learning results achieved by training the model on two separate source domains, namely CBF and CinCECGTorso in the UCR archive. We then evaluate the model's performance on the downstream classification task across other $9$ target domains in the first $10$ datasets in the UCR archive. The average results are 0.864 for CBF and 0.895 for CinCECGTorso and 0.912 for no transfer scenario. The transformation results show competitive performance with no transfer scenario. More transfer learning results are in the Appendix.

\begin{table}[h]
  \centering
  \resizebox{0.87\linewidth}{!}{
  \begin{tabular}{lcccccccc}
  \toprule
    & \multicolumn{3}{c}{Yahoo} & \multicolumn{3}{c}{KPI} \\
    \cmidrule(r){2-4} \cmidrule(r){5-7}
    & F$_1$ & Prec. & Rec. & F$_1$ & Prec. & Rec. \\
    \midrule
    SPOT & 0.338 & 0.269 & 0.454 & 0.217 & 0.786 & 0.126\\
    DSPOT & 0.316 & 0.241 & 0.458 & 0.521 & 0.623 & 0.447 \\
    DONUT & 0.026 & 0.013 & 0.825 & 0.347 & 0.371 & 0.326 \\
    SR & 0.563 & 0.451 & 0.747 & 0.622 & 0.647 & 0.598\\
    TS2Vec & 0.745 & 0.729 & 0.762 & 0.677 & 0.929 & 0.533 \\
    TimesURL & \textbf{0.749} & 0.748 &  0.750 & \textbf{0.688} & 0.925 & 0.546 \\
    \bottomrule
  \end{tabular}}
  \caption{Univariate time series anomaly detection results.}
  \label{tab:anomaly}
\end{table}

\begin{table}[h]
        \centering
        \resizebox{0.69\linewidth}{!}{
        \begin{tabular}{c c}
            \\
            \toprule
            & Avg. Accuracy \\
            \midrule
            \textbf{TimesURL} &  \textbf{0.752} \\
            w/o Frequency Mixing  &  0.709 (-4.3\%) \\
            w/o Instance Universum  & 0.720 (-3.2\%) \\
            w/o Temporal Universum  & 0.717 (-3.5\%) \\
            w/o Double Universum  & 0.716 (-3.6\%) \\
            w/o Time Reconstruction & 0.735 (-1.8\%) \\
            \bottomrule
        \end{tabular}
        }
    \caption{Ablation results on 30 UEA datasets.} \label{tab:ablation}
\end{table}

\subsection{Ablation Study}
Throughout the paper, we emphasize the importance of suitable augmentation methods, enough hard negative samples and proper both segment- and instance-levels of information for learning universal representations and design FTAug, Double Universums and joint optimization strategies to satisfy the above requirements respectively. To verify the effectiveness of the above three modules in TimesURL, a comparison between full TimesURL and its five variants on 30 datasets in the UEA archive is shown in Table \ref{tab:ablation}, where 1) w/o frequency mixing, 2) w/o instance Universum, 3) w/o temporal Universum, 4) w/o double Universum, 5) w/o time reconstruction. Results show that all the above components of TimesURL are indispensable. We have to mention that constructing either temporal- or instance-wise Universums cannot achieve optimal performance, while double Universums achieve better performance by providing sufficient and discriminative information for both temporal- and instance-wise contrastive learning.



\section{Conclusion}
In this paper, we propose a novel self-supervised framework termed TimesURL that can learn universal time series representations for various types of downstream tasks.   
We introduce a new augmentation method called FTAug to keep contextual consistency and temporal characteristics unchanged, which is suitable for various downstream tasks. Moreover, we inject double Universums into contrastive learning to enhance negative sample quantity and quality to boost the performance of contrastive learning. Last but not least, TimesURL jointly optimizes contrastive learning and time reconstruction for capturing both segment- and instance-levels of information for universal representation learning. Experimental results demonstrate the effectiveness of the above strategies and show that with suitable augmentation methods, enough hard negative samples and proper levels of information, TimesURL shows great performance on six downstream tasks.

\section{Acknowledgments}
The authors express their gratitude to Meng Cao for assisting with the implementation of the code, Xiang Li and Jiaqiang Zhang for proofreading this manuscript, and Zhengyu Cao for providing the datasets. This work is supported by the Key Program of NSFC under Grant No.62076124, Postgraduate Research \& Practice Innovation Program of Jiangsu Province under Grant No.KYCX21\_0225 and Scientific and Technological Achievements Transferring Project of Jiangsu Province under Grant No. BA2021005.

\bibliography{aaai24}

\begin{thebibliography}{48}
\providecommand{\natexlab}[1]{#1}

\bibitem[{Bagnall et~al.(2018)Bagnall, Dau, Lines, Flynn, Large, Bostrom,
  Southam, and Keogh}]{bagnall2018uea}
Bagnall, A.; Dau, H.~A.; Lines, J.; Flynn, M.; Large, J.; Bostrom, A.; Southam,
  P.; and Keogh, E. 2018.
\newblock The UEA multivariate time series classification archive, 2018.
\newblock \emph{arXiv preprint arXiv:1811.00075}.

\bibitem[{Bai, Kolter, and Koltun(2018)}]{bai2018empirical}
Bai, S.; Kolter, J.~Z.; and Koltun, V. 2018.
\newblock An empirical evaluation of generic convolutional and recurrent
  networks for sequence modeling.
\newblock \emph{arXiv preprint arXiv:1803.01271}.

\bibitem[{Bayer et~al.(2020)Bayer, Soelch, Mirchev, Kayalibay, and van~der
  Smagt}]{bayer2020mind}
Bayer, J.; Soelch, M.; Mirchev, A.; Kayalibay, B.; and van~der Smagt, P. 2020.
\newblock Mind the Gap when Conditioning Amortised Inference in Sequential
  Latent-Variable Models.
\newblock In \emph{International Conference on Learning Representations}.

\bibitem[{Cai et~al.(2020)Cai, Frankle, Schwab, and Morcos}]{Cai2020AreAN}
Cai, T.; Frankle, J.; Schwab, D.~J.; and Morcos, A.~S. 2020.
\newblock Are all negatives created equal in contrastive instance
  discrimination?
\newblock \emph{ArXiv}, abs/2010.06682.

\bibitem[{Chapelle et~al.(2007)Chapelle, Agarwal, Sinz, and
  Sch{\"o}lkopf}]{chapelle2007analysis}
Chapelle, O.; Agarwal, A.; Sinz, F.; and Sch{\"o}lkopf, B. 2007.
\newblock An analysis of inference with the universum.
\newblock \emph{Advances in neural information processing systems}, 20.

\bibitem[{Chen et~al.(2023)Chen, Xu, Zeng, and Xu}]{chen2023fraug}
Chen, M.; Xu, Z.; Zeng, A.; and Xu, Q. 2023.
\newblock FrAug: Frequency Domain Augmentation for Time Series Forecasting.
\newblock \emph{arXiv preprint arXiv:2302.09292}.

\bibitem[{Chen et~al.(2020)Chen, Kornblith, Norouzi, and
  Hinton}]{chen2020simple}
Chen, T.; Kornblith, S.; Norouzi, M.; and Hinton, G. 2020.
\newblock A simple framework for contrastive learning of visual
  representations.
\newblock In \emph{International conference on machine learning}, 1597--1607.
  PMLR.

\bibitem[{Chung et~al.(2015)Chung, Kastner, Dinh, Goel, Courville, and
  Bengio}]{chung2015recurrent}
Chung, J.; Kastner, K.; Dinh, L.; Goel, K.; Courville, A.~C.; and Bengio, Y.
  2015.
\newblock A recurrent latent variable model for sequential data.
\newblock \emph{Advances in neural information processing systems}, 28.

\bibitem[{Dau et~al.(2019)Dau, Bagnall, Kamgar, Yeh, Zhu, Gharghabi,
  Ratanamahatana, and Keogh}]{dau2019ucr}
Dau, H.~A.; Bagnall, A.; Kamgar, K.; Yeh, C.-C.~M.; Zhu, Y.; Gharghabi, S.;
  Ratanamahatana, C.~A.; and Keogh, E. 2019.
\newblock The UCR time series archive.
\newblock \emph{IEEE/CAA Journal of Automatica Sinica}, 6(6): 1293--1305.

\bibitem[{Dempster, Petitjean, and Webb(2020)}]{dempster2020rocket}
Dempster, A.; Petitjean, F.; and Webb, G.~I. 2020.
\newblock ROCKET: exceptionally fast and accurate time series classification
  using random convolutional kernels.
\newblock \emph{Data Mining and Knowledge Discovery}, 34(5): 1454--1495.

\bibitem[{Denton et~al.(2017)}]{denton2017unsupervised}
Denton, E.~L.; et~al. 2017.
\newblock Unsupervised learning of disentangled representations from video.
\newblock \emph{Advances in neural information processing systems}, 30.

\bibitem[{Eldele et~al.(2021)Eldele, Ragab, Chen, Wu, Kwoh, Li, and
  Guan}]{Eldele2021TimeSeriesRL}
Eldele, E.; Ragab, M.; Chen, Z.; Wu, M.; Kwoh, C.; Li, X.; and Guan, C. 2021.
\newblock Time-Series Representation Learning via Temporal and Contextual
  Contrasting.
\newblock In \emph{International Joint Conference on Artificial Intelligence}.

\bibitem[{Eldele et~al.(2022)Eldele, Ragab, Chen, Wu, Kwoh, Li, and
  Guan}]{emadeldeen2022catcc}
Eldele, E.; Ragab, M.; Chen, Z.; Wu, M.; Kwoh, C.~K.; Li, X.; and Guan, C.
  2022.
\newblock Self-supervised Contrastive Representation Learning for
  Semi-supervised Time-Series Classification.
\newblock \emph{arXiv preprint arXiv:2208.06616}.

\bibitem[{Grill et~al.(2020)Grill, Strub, Altch{\'e}, Tallec, Richemond,
  Buchatskaya, Doersch, Avila~Pires, Guo, Gheshlaghi~Azar
  et~al.}]{grill2020bootstrap}
Grill, J.-B.; Strub, F.; Altch{\'e}, F.; Tallec, C.; Richemond, P.;
  Buchatskaya, E.; Doersch, C.; Avila~Pires, B.; Guo, Z.; Gheshlaghi~Azar, M.;
  et~al. 2020.
\newblock Bootstrap your own latent-a new approach to self-supervised learning.
\newblock \emph{Advances in neural information processing systems}, 33:
  21271--21284.

\bibitem[{Gutmann and Hyv{\"a}rinen(2012)}]{gutmann2012noise}
Gutmann, M.~U.; and Hyv{\"a}rinen, A. 2012.
\newblock Noise-Contrastive Estimation of Unnormalized Statistical Models, with
  Applications to Natural Image Statistics.
\newblock \emph{Journal of machine learning research}, 13(2).

\bibitem[{Han and Chen(2023)}]{han2023universum}
Han, A.; and Chen, S. 2023.
\newblock Universum-Inspired Supervised Contrastive Learning.
\newblock In \emph{Web and Big Data: 6th International Joint Conference,
  APWeb-WAIM 2022, Nanjing, China, November 25--27, 2022, Proceedings, Part
  II}, 459--473. Springer.

\bibitem[{He et~al.(2022)He, Chen, Xie, Li, Doll{\'a}r, and
  Girshick}]{he2022masked}
He, K.; Chen, X.; Xie, S.; Li, Y.; Doll{\'a}r, P.; and Girshick, R. 2022.
\newblock Masked autoencoders are scalable vision learners.
\newblock In \emph{Proceedings of the IEEE/CVF Conference on Computer Vision
  and Pattern Recognition}, 16000--16009.

\bibitem[{Kalantidis et~al.(2020)Kalantidis, Sariyildiz, Pion, Weinzaepfel, and
  Larlus}]{kalantidis2020hard}
Kalantidis, Y.; Sariyildiz, M.~B.; Pion, N.; Weinzaepfel, P.; and Larlus, D.
  2020.
\newblock Hard negative mixing for contrastive learning.
\newblock \emph{Advances in Neural Information Processing Systems}, 33:
  21798--21809.

\bibitem[{Kenton and Toutanova(2019)}]{kenton2019bert}
Kenton, J. D. M.-W.~C.; and Toutanova, L.~K. 2019.
\newblock BERT: Pre-training of Deep Bidirectional Transformers for Language
  Understanding.
\newblock In \emph{Proceedings of NAACL-HLT}, 4171--4186.

\bibitem[{Krishnan, Shalit, and Sontag(2017)}]{krishnan2017structured}
Krishnan, R.; Shalit, U.; and Sontag, D. 2017.
\newblock Structured inference networks for nonlinear state space models.
\newblock In \emph{Proceedings of the AAAI Conference on Artificial
  Intelligence}, volume~31.

\bibitem[{Lei et~al.(2019)Lei, Yi, Vaculin, Wu, and
  Dhillon}]{lei2017similarity}
Lei, Q.; Yi, J.; Vaculin, R.; Wu, L.; and Dhillon, I.~S. 2019.
\newblock Similarity Preserving Representation Learning for Time Series
  Clustering.
\newblock In \emph{Proceedings of the 28th International Joint Conference on
  Artificial Intelligence}, IJCAI'19, 2845–2851. AAAI Press.
\newblock ISBN 9780999241141.

\bibitem[{Li et~al.(2019)Li, Jin, Xuan, Zhou, Chen, Wang, and
  Yan}]{li2019enhancing}
Li, S.; Jin, X.; Xuan, Y.; Zhou, X.; Chen, W.; Wang, Y.-X.; and Yan, X. 2019.
\newblock Enhancing the locality and breaking the memory bottleneck of
  transformer on time series forecasting.
\newblock \emph{Advances in neural information processing systems}, 32.

\bibitem[{Liu and Chen(2019)}]{liu2019non}
Liu, J.; and Chen, S. 2019.
\newblock Non-stationary multivariate time series prediction with selective
  recurrent neural networks.
\newblock In \emph{Pacific rim international conference on artificial
  intelligence}, 636--649. Springer.

\bibitem[{Liu et~al.(2022{\natexlab{a}})Liu, Zeng, Chen, Xu, Lai, Ma, and
  Xu}]{liu2022scinet}
Liu, M.; Zeng, A.; Chen, M.; Xu, Z.; Lai, Q.; Ma, L.; and Xu, Q.
  2022{\natexlab{a}}.
\newblock Scinet: Time series modeling and forecasting with sample convolution
  and interaction.
\newblock \emph{Advances in Neural Information Processing Systems}, 35:
  5816--5828.

\bibitem[{Liu and wei Liu(2022)}]{Liu2022TheTP}
Liu, Y.; and wei Liu, J. 2022.
\newblock The Time-Sequence Prediction via Temporal and Contextual Contrastive
  Representation Learning.
\newblock In \emph{Pacific Rim International Conference on Artificial
  Intelligence}.

\bibitem[{Liu et~al.(2022{\natexlab{b}})Liu, Wu, Wang, and Long}]{liu2022non}
Liu, Y.; Wu, H.; Wang, J.; and Long, M. 2022{\natexlab{b}}.
\newblock Non-stationary Transformers: Exploring the Stationarity in Time
  Series Forecasting.
\newblock In \emph{Advances in Neural Information Processing Systems}.

\bibitem[{Luo et~al.(2023)Luo, Cheng, Wang, Xu, Ni, Yu, Zhang, Liu, Chen, Chen
  et~al.}]{luo2023time}
Luo, D.; Cheng, W.; Wang, Y.; Xu, D.; Ni, J.; Yu, W.; Zhang, X.; Liu, Y.; Chen,
  Y.; Chen, H.; et~al. 2023.
\newblock Time Series Contrastive Learning with Information-Aware
  Augmentations.
\newblock In \emph{Proceedings of the AAAI Conference on Artificial
  Intelligence}.

\bibitem[{Ma et~al.(2019)Ma, Zheng, Li, and Cottrell}]{ma2019learning}
Ma, Q.; Zheng, J.; Li, S.; and Cottrell, G.~W. 2019.
\newblock Learning representations for time series clustering.
\newblock \emph{Advances in neural information processing systems}, 32.

\bibitem[{Malhotra et~al.(2017)Malhotra, TV, Vig, Agarwal, and
  Shroff}]{malhotra2017timenet}
Malhotra, P.; TV, V.; Vig, L.; Agarwal, P.; and Shroff, G. 2017.
\newblock TimeNet: Pre-trained deep recurrent neural network for time series
  classification.
\newblock \emph{arXiv preprint arXiv:1706.08838}.

\bibitem[{Nikolay~Laptev(2015)}]{yahoo}
Nikolay~Laptev, Y.~B., Saeed~Amizadeh. 2015.
\newblock A Benchmark Dataset for Time Series Anomaly Detection.
\newblock
  \url{https://yahooresearch.tumblr.com/post/114590420346/a-benchmark-dataset-for-time-series-anomaly}.

\bibitem[{Oreshkin et~al.(2019)Oreshkin, Carpov, Chapados, and
  Bengio}]{oreshkin2019n}
Oreshkin, B.~N.; Carpov, D.; Chapados, N.; and Bengio, Y. 2019.
\newblock N-BEATS: Neural basis expansion analysis for interpretable time
  series forecasting.
\newblock \emph{arXiv preprint arXiv:1905.10437}.

\bibitem[{Pagliardini, Gupta, and Jaggi(2017)}]{pagliardini2017unsupervised}
Pagliardini, M.; Gupta, P.; and Jaggi, M. 2017.
\newblock Unsupervised learning of sentence embeddings using compositional
  n-gram features.
\newblock \emph{arXiv preprint arXiv:1703.02507}.

\bibitem[{Ren et~al.(2019)Ren, Xu, Wang, Yi, Huang, Kou, Xing, Yang, Tong, and
  Zhang}]{ren2019time}
Ren, H.; Xu, B.; Wang, Y.; Yi, C.; Huang, C.; Kou, X.; Xing, T.; Yang, M.;
  Tong, J.; and Zhang, Q. 2019.
\newblock Time-series anomaly detection service at microsoft.
\newblock In \emph{Proceedings of the 25th ACM SIGKDD international conference
  on knowledge discovery \& data mining}, 3009--3017.

\bibitem[{Robinson et~al.(2020)Robinson, Chuang, Sra, and
  Jegelka}]{robinson2020contrastive}
Robinson, J.; Chuang, C.-Y.; Sra, S.; and Jegelka, S. 2020.
\newblock Contrastive learning with hard negative samples.
\newblock \emph{arXiv preprint arXiv:2010.04592}.

\bibitem[{Shi et~al.(2015)Shi, Chen, Wang, Yeung, Wong, and
  Woo}]{shi2015convolutional}
Shi, X.; Chen, Z.; Wang, H.; Yeung, D.-Y.; Wong, W.-K.; and Woo, W.-c. 2015.
\newblock Convolutional LSTM network: A machine learning approach for
  precipitation nowcasting.
\newblock \emph{Advances in neural information processing systems}, 28.

\bibitem[{Siffer et~al.(2017)Siffer, Fouque, Termier, and
  Largouet}]{siffer2017anomaly}
Siffer, A.; Fouque, P.-A.; Termier, A.; and Largouet, C. 2017.
\newblock Anomaly detection in streams with extreme value theory.
\newblock In \emph{Proceedings of the 23rd ACM SIGKDD International Conference
  on Knowledge Discovery and Data Mining}, 1067--1075.

\bibitem[{Tonekaboni, Eytan, and Goldenberg(2021)}]{tonekaboni2021unsupervised}
Tonekaboni, S.; Eytan, D.; and Goldenberg, A. 2021.
\newblock Unsupervised Representation Learning for Time Series with Temporal
  Neighborhood Coding.
\newblock In \emph{International Conference on Learning Representations}.

\bibitem[{Um et~al.(2017)Um, Pfister, Pichler, Endo, Lang, Hirche, Fietzek, and
  Kuli{\'c}}]{um2017data}
Um, T.~T.; Pfister, F.~M.; Pichler, D.; Endo, S.; Lang, M.; Hirche, S.;
  Fietzek, U.; and Kuli{\'c}, D. 2017.
\newblock Data augmentation of wearable sensor data for parkinson’s disease
  monitoring using convolutional neural networks.
\newblock In \emph{Proceedings of the 19th ACM international conference on
  multimodal interaction}, 216--220.

\bibitem[{Vapnik(2006)}]{vapnik2006Transductive}
Vapnik, V. 2006.
\newblock Transductive Inference and Semi-Supervised Learning.
\newblock \emph{Semi-Supervised Learning}, 453--472.

\bibitem[{Wang and Gupta(2015)}]{wang2015unsupervised}
Wang, X.; and Gupta, A. 2015.
\newblock Unsupervised learning of visual representations using videos.
\newblock In \emph{Proceedings of the IEEE international conference on computer
  vision}, 2794--2802.

\bibitem[{Woo et~al.(2022)Woo, Liu, Sahoo, Kumar, and Hoi}]{woo2022cost}
Woo, G.; Liu, C.; Sahoo, D.; Kumar, A.; and Hoi, S. 2022.
\newblock CoST: Contrastive Learning of Disentangled Seasonal-Trend
  Representations for Time Series Forecasting.
\newblock In \emph{International Conference on Learning Representations}.

\bibitem[{Wu et~al.(2023)Wu, Hu, Liu, Zhou, Wang, and Long}]{wu2023timesnet}
Wu, H.; Hu, T.; Liu, Y.; Zhou, H.; Wang, J.; and Long, M. 2023.
\newblock TimesNet: Temporal 2D-Variation Modeling for General Time Series
  Analysis.
\newblock In \emph{The Eleventh International Conference on Learning
  Representations}.

\bibitem[{Wu et~al.(2021)Wu, Xu, Wang, and Long}]{wu2021autoformer}
Wu, H.; Xu, J.; Wang, J.; and Long, M. 2021.
\newblock Autoformer: Decomposition transformers with auto-correlation for
  long-term series forecasting.
\newblock \emph{Advances in Neural Information Processing Systems}, 34:
  22419--22430.

\bibitem[{Xu et~al.(2018)Xu, Chen, Zhao, Li, Bu, Li, Liu, Zhao, Pei, Feng
  et~al.}]{xu2018unsupervised}
Xu, H.; Chen, W.; Zhao, N.; Li, Z.; Bu, J.; Li, Z.; Liu, Y.; Zhao, Y.; Pei, D.;
  Feng, Y.; et~al. 2018.
\newblock Unsupervised anomaly detection via variational auto-encoder for
  seasonal kpis in web applications.
\newblock In \emph{Proceedings of the 2018 world wide web conference},
  187--196.

\bibitem[{Xu et~al.(2022)Xu, Lian, Zhao, Gong, Shou, Jiang, Xie, and
  Wen}]{xu2022negative}
Xu, L.; Lian, J.; Zhao, W.~X.; Gong, M.; Shou, L.; Jiang, D.; Xie, X.; and Wen,
  J.-R. 2022.
\newblock Negative sampling for contrastive representation learning: A review.
\newblock \emph{arXiv preprint arXiv:2206.00212}.

\bibitem[{Yue et~al.(2022)Yue, Wang, Duan, Yang, Huang, Tong, and
  Xu}]{yue2022ts2vec}
Yue, Z.; Wang, Y.; Duan, J.; Yang, T.; Huang, C.; Tong, Y.; and Xu, B. 2022.
\newblock Ts2vec: Towards universal representation of time series.
\newblock In \emph{Proceedings of the AAAI Conference on Artificial
  Intelligence}, volume~36, 8980--8987.

\bibitem[{Zerveas et~al.(2021)Zerveas, Jayaraman, Patel, Bhamidipaty, and
  Eickhoff}]{zerveas2021transformer}
Zerveas, G.; Jayaraman, S.; Patel, D.; Bhamidipaty, A.; and Eickhoff, C. 2021.
\newblock A transformer-based framework for multivariate time series
  representation learning.
\newblock In \emph{Proceedings of the 27th ACM SIGKDD Conference on Knowledge
  Discovery \& Data Mining}, 2114--2124.

\bibitem[{Zhou et~al.(2021)Zhou, Zhang, Peng, Zhang, Li, Xiong, and
  Zhang}]{zhou2021informer}
Zhou, H.; Zhang, S.; Peng, J.; Zhang, S.; Li, J.; Xiong, H.; and Zhang, W.
  2021.
\newblock Informer: Beyond efficient transformer for long sequence time-series
  forecasting.
\newblock In \emph{Proceedings of the AAAI conference on artificial
  intelligence}, volume~35, 11106--11115.

\end{thebibliography}

\end{document}